%% file: egpaper_for_review.tex
\PassOptionsToPackage{dvipsnames}{xcolor}
\documentclass[10pt,twocolumn,letterpaper]{article}

\usepackage{iccv}
\usepackage{times}

\usepackage[sort,compress]{cite}

\usepackage{graphicx}
\usepackage[dvipsnames]{xcolor}

\usepackage{amsmath}
\usepackage{amssymb}

\usepackage{fontawesome5}

\renewcommand{\paragraph}[1]{\noindent\textbf{#1}}

\usepackage{subcaption}
\usepackage[inline]{enumitem}
\usepackage{tikz}
\usetikzlibrary{
    calc,
    matrix,
    positioning,
}

\usepackage{pgfplots}
\usetikzlibrary{patterns}
\pgfplotsset{compat=1.18}
\usepgfplotslibrary{
    colorbrewer,
    fillbetween,
    groupplots,
}

\pgfplotsset{%
  compat=1.18,
  grid style=dashed,
  ymajorgrids=true,
  cycle list/Dark2,
  dark marker list/.style={cycle multiindex* list={
    Dark2\nextlist
    mark list\nextlist
    mark options={fill=.!75}
  }},
  dark list/.style={cycle multiindex* list={
    Dark2\nextlist
    fill=.!75
  }},
}

\usepackage{booktabs}

\usepackage{siunitx}
\robustify\textbf
\robustify\bfseries

\usepackage{multirow}

\usepackage{makecell}

\usepackage{amsmath}
\usepackage{amsthm}
\usepackage{commath}

\DeclareMathOperator*{\argmin}{arg\,min}

\let\given\givenbase

\DeclareMathOperator{\logit}{logit}

\usepackage{dutchcal}

\usepackage{algorithm}
\usepackage{algorithmic}

\usepackage{xspace}
\usepackage{centernot}
\makeatletter
\DeclareRobustCommand\onedot{\futurelet\@let@token\@onedot}
\def\@onedot{\ifx\@let@token.\else.\null\fi\xspace}

\def\eg{{e.g}\onedot} 
\def\ie{{i.e}\onedot} 
\def\cf{{cf}\onedot}

\makeatother

\newcommand\nomarkfootnote[1]{%
  \begingroup
  \renewcommand\thefootnote{}\footnote{#1}%
  \addtocounter{footnote}{-1}%
  \endgroup
}

\usepackage[pagebackref=true,breaklinks=true,letterpaper=true,colorlinks,bookmarks=false]{hyperref}
\hypersetup{
  colorlinks,
  linkcolor = BrickRed,
  citecolor = RoyalBlue,
  urlcolor  = WildStrawberry,
}
\iccvfinalcopy %

\def\iccvPaperID{****} %

\ificcvfinal\fi

\begin{document}

\title{SelfGraphVQA: A Self-Supervised Graph Neural Network for Scene-based Question Answering}

\author{Bruno Souza\thanks{Work carried out as Guest Researcher at UiO.}\\
{\small University of Campinas}\\
{\tt\small b234837@dac.unicamp.br}
\and
Marius Aasan\\
{\small University of Oslo}\\
{\tt\small mariuaas@uio.no}
\and
Helio Pedrini\\
{\small University of Campinas}\\
{\tt\small helio@ic.unicamp.br}
\and
Ad\'in Ram\'irez Rivera\\
{\small University of Oslo}\\
{\tt\small adinr@uio.no}
}

\maketitle
\nomarkfootnote{To appear in Vision-and-Language Algorithmic Reasoning Workshop at ICCV 2023}

\ificcvfinal\thispagestyle{empty}\fi
\vspace*{-10pt}

\begin{abstract}
\noindent The intersection of vision and language is of major interest due to the increased focus on seamless integration between recognition and reasoning.
Scene graphs (SGs) have emerged as a useful tool for multimodal image analysis, showing impressive performance in tasks such as Visual Question Answering (VQA).
In this work, we demonstrate that despite the effectiveness of scene graphs in VQA tasks, current methods that utilize idealized annotated scene graphs struggle to generalize when using predicted scene graphs extracted from images.
To address this issue, we introduce the SelfGraphVQA framework.
Our approach extracts a scene graph from an input image using a pre-trained scene graph generator and employs semantically-preserving augmentation with self-supervised techniques.
This method improves the utilization of graph representations in VQA tasks by circumventing the need for costly and potentially biased annotated data.
By creating alternative views of the extracted graphs through image augmentations, we can learn joint embeddings by optimizing the informational content in their representations using an un-normalized contrastive approach.
As we work with SGs, we experiment with three distinct maximization strategies: node-wise, graph-wise, and permutation-equivariant regularization.
We empirically showcase the effectiveness of the extracted scene graph for VQA and demonstrate that these approaches enhance overall performance by highlighting the significance of visual information.
This offers a more practical solution for VQA tasks that rely on SGs for complex reasoning questions.
\end{abstract}

\section{Introduction}

The successful execution of Visual Question Answering (VQA) relies on a comprehensive understanding of the scene, including spatial interrelationships and reasoning inference capabilities~\cite{agrawal2018don,gva2019dataset}.
Incorporating scene graph (SG) representations in SG-VQA tasks has shown promising outcomes~\cite{knyazev2020graph,hu2019language,li2019relationaware,wang2022sgeitl,nuthalapati2021lightweight}, providing concise representations of complex spatial and relational information.

Earlier investigations into SG-VQA demonstrated that successful models primarily rely on the utilization of manually annotated scene graphs for training~\cite{liang2021graphvqa, liang2020lrta, nuthalapati2021lightweight}, resulting in remarkably high levels of accuracy on the GQA dataset~\cite{gva2019dataset}, surpassing human performance by a significant margin (see Table~\ref{tab:previous}).

Despite the promising results, we argue that utilizing pre-annotated SGs in VQA is impractical in the real world due to its labor-intensive nature.
Also, it permits a wide range of semantically corresponding SG~\cite{he2020scene} and when annotated it could potentially introduce questions-related biases, giving rise to concerns about its generalizability~\cite{agrawal2023reassessing}.
These issues may limit the model's ability to solve real-world problems beyond the dataset~\cite{luo2021just}.
This is evident in a significant decline in accuracy, approximately 60\% when models are confronted with automatically generated SGs\@.
Additionally, studies assert that the main limitation in generalizing stems largely from linguistic correlations.~\cite{agrawal2023reassessing,lake2018generalization}.

\begin{figure}[tb]
\centering
\resizebox{\linewidth}{!}{%
\begin{tikzpicture}[
    node distance=.5cm,
    var/.style={
        draw,
        circle,
        minimum width=0.6cm,
        inner sep=2pt,
        fill=#1,
        font=\small,
    },
    conn/.style={
        ->,
        shorten <=2pt,
        shorten >=2pt,
    },
    proc/.style={
        draw,
        rectangle,
        rounded corners,
        text width=1.1cm,
        minimum height=.75cm,
        align=center,
        font=\large,
    },
    graph/.style={
        matrix of nodes,
        row sep=10pt, column sep=10pt,
        nodes={draw, circle, font=\tiny, text width=.6cm, align=center, inner sep=0pt, fill={rgb:orange,3;white,1;pink,1}},
    },
    graph2/.style={
        matrix of nodes,
        row sep=10pt, column sep=10pt,
        nodes={draw, circle, font=\tiny, text width=.6cm, align=center, inner sep=0pt, fill={rgb:orange,1;white,1;pink,1}},
    },
    edge/.style={
        shorten <=2pt,
        shorten >=2pt,
        ->,
        rounded corners,
    },
    loss/.style={
        draw,
        circle,
        font=\scriptsize,
    },
]
\begin{scope}[node distance=0.75cm]
\node[var=black!25, text=black] (g) {G};
\node[var=white, text=black, above right=of g] (q) {Q};
\node[var=white, text=black, below right=of g] (i) {I};
\node[var=white, text=black, right=of $(q)!.5!(i)$] (a) {A};
\draw[conn] (g) -- (q);
\draw[conn] (g) -- (i);
\draw[conn, dashed] (i) -- (a);
\draw[conn, dashed] (q) -- (a);
\draw[conn, dotted, orange] (g) -- node[midway, font=\small, inner sep=0pt, text width=1.4cm, align=center] {Spurious\\correlation} (a);
\draw[conn, dotted, orange] (i) -- (q);

\node[var=black!25, text=black, below=1.5cm of g |- i] (g2) {G};
\node[var=white, text=black, above right=of g2] (q2) {Q};
\node[var=white, text=black, below right=of g2] (i2) {I};
\node[var=LimeGreen!50, text=black] (gp) at ($(q2)!.5!(i2)$) {G'};
\node[var=white, text=black, right=of $(q2)!.5!(i2)$] (a2) {A};
\draw[conn, black!25] (g2) -- node[midway, red!50] {\small\faIcon{times}} (q2);
\draw[conn, black!25] (g2) -- node[midway, red!50] {\small\faIcon{times}} (i2);
\draw[conn] (i2) -- (gp);
\draw[conn, dashed] (gp) -- (a2);
\draw[conn, dashed] (q2) -- (a2);
\end{scope}

\node[below right=.1cm and .5cm of a] (io) {\includegraphics[width=1.8cm]{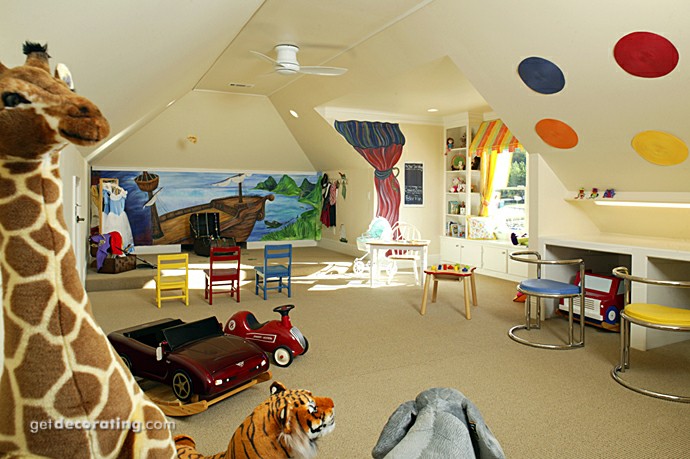}};
\node[below=.75cm of io] (it) {\includegraphics[width=1.8cm]{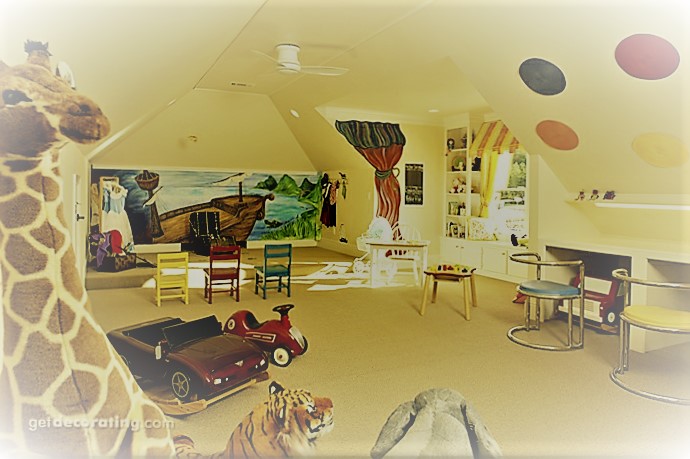}};

\node[above=.75cm of io, rectangle, draw, font=\scriptsize, text width=1.8cm] (txt) {How many chairs are in the room?};

\node[right=of txt, proc, fill=white!70!NavyBlue] (enc-q) {\large $f_q$};

\node[right=of io, proc, fill=white!50!LimeGreen] (sgg-o) {\large $g$};
\node[right=of it, proc, fill=white!50!LimeGreen] (sgg-t) {\large $g$};

\matrix (go) [right=of sgg-o, graph, ampersand replacement=\&,] {
Wall \& Chair \\
Table \& Toy \\
};
\draw[edge] (go-1-1) -- (go-1-2);
\draw[edge] (go-2-1) -- (go-1-1);
\draw[edge] (go-2-2) -- (go-1-2);

\matrix (gt) [right=of sgg-t, graph2, ampersand replacement=\&] {
Chair \& Toy \\
Balls \& Floor \\
};
\draw[edge] (gt-1-2) -- (gt-1-1);
\draw[edge] (gt-1-1) -- (gt-2-1);
\draw[edge] (gt-2-2) -- (gt-1-2);

\node[right=of go, proc, fill=white!70!NavyBlue] (enc-o) {\large $f_g$};
\node[right=of gt, proc, fill=white!70!NavyBlue] (enc-t) {\large $f_g$};

\node[right=of enc-o, proc, fill=white!75!black] (proj) {\large $h$};
\node[above=of proj, proc, fill=white!70!NavyBlue] (clas) {\large $f_c$};

\coordinate (enc-mid) at ({$(enc-o)!.5!(enc-t)$} -| proj.east);
\node[right=of enc-mid, loss, fill=white!60!red] (l-sim) {\large $L'$};

\node[right=of clas, loss, fill=white!60!red] (l-ans) {\large $L_\mathcal{a}$};

\draw[edge] (io) -- (sgg-o);
\draw[edge] (sgg-o) -- (go);
\draw[edge] (txt) -- (enc-q);

\draw[edge] (it) -- (sgg-t);
\draw[edge] (sgg-t) -- (gt);

\coordinate (enc-o-1) at ($(enc-o.north west)!.25!(enc-o.south west)$);
\coordinate (enc-o-2) at ($(enc-o.north west)!.75!(enc-o.south west)$);
\draw[edge] (go.east |- enc-o-2) -- (enc-o-2);

\coordinate (enc-t-1) at ($(enc-t.north west)!.25!(enc-t.south west)$);
\coordinate (enc-t-2) at ($(enc-t.north west)!.75!(enc-t.south west)$);
\draw[edge] (gt.east |- enc-t-2) -- (enc-t-2);

\draw[edge] (enc-q) -| ($(enc-o-1)+(-.3cm,0)$) |- (enc-o-1);
\draw[edge] ($(enc-o-1)+(-.3cm,5pt)$) |- (enc-t-1);

\draw[edge] (enc-o) -- node[midway, font=\small, inner sep=2pt, above] {\large $z_1$}  (proj);

\draw[edge] (proj) -| (l-sim);
\draw[edge, dashed]  (enc-t)  -|  (l-sim);

\path (proj) -- node[midway, font=\small, inner sep=2pt, above] {\large $p_1$} (proj -| l-sim);
\path (enc-t) -- node[midway, font=\small, inner sep=2pt, above] {\large $z_2$} (enc-t -| l-sim);

\draw[edge] (enc-o) |- (clas);
\draw[edge] (clas) -- (l-ans);

\draw[edge, <->, dashed] (sgg-o) -- (sgg-t) node[midway, font=\scriptsize, fill=white, inner sep=1pt] {Shared + Frozen};

\draw[edge, <->, dashed] (enc-o) -- (enc-t) node[midway, font=\scriptsize, fill=white, inner sep=1pt] {Shared};

\node[above=.5cm of l-sim, fill=white] (grad) {$\nabla$};
\def\notnabla{\mathrel{\ooalign{\raisebox{1pt}{$\Large{\not}$}\hss\cr$\nabla$}}}
\node[below=.5cm of l-sim, fill=white] (ngrad) {$\notnabla$};

\end{tikzpicture}}
    \caption{(Left) The statistical dependence of the task and the ideal graph, $G$. (Right) Our proposed framework removes data leakage by using the extracted SG $G'$. Our architecture comprises a question encoder $f_q$, a graph encoder $f_g$, and a classifier $f_c$. Two distinct views of one image are processed by the same pipeline. We use a frozen pre-trained SG generator $g$, and a prediction head $h$ is applied through the top view with gradient backpropagation, while gradients are not propagated back from the lower view. We maximize the representation of the views using the similarity loss $L'$.}
    \label{fig:statisticdependence}
    \label{fig:SelfGraphVQA}
\end{figure}
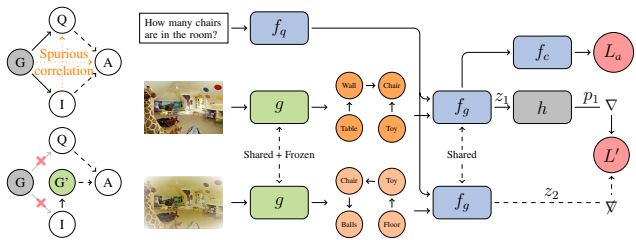

In this study, we address these challenges by extracting an SG from a given image using an unbiased, off-the-shelf scene graph generator~\cite{knyazev2020graph}, with the aim of removing any potential information leakage, as illustrated in Fig.~\ref{fig:statisticdependence}'s structure.
Furthermore, our method employs semantically preserving augmentation, integrated with un-normalized contrastive framework, to further mitigate potential linguistic biases to enhance the visual cues translated as SG for VQA\@.
We refer to it as the \textit{SelfGraphVQA framework}, \cf Fig.~\ref{fig:statisticdependence}.

Given its simplicity~\cite{chen2021exploring}, our approach is trained using joint embeddings and a Siamese network architecture, inspired by the SimSiam model, which does not require negative samples~\cite{bromley1993signature, garrido2022duality}.
In this work, we explore three un-normalized contrastive approaches (node-wise, graph-wise, and regularization for permutation equivariance) and demonstrate its effectiveness by enhancing the visual information for the VQA task.
A graph neural network (GNN) with a self-attention strategy (GAT) is employed to distill an SG representation relevant to the question by capturing visual interaction content among objects in the scene~\cite{chen2021exploring}.

Our work differs from existing VQA models in three main aspects: (i)~we generate as SG using a pre-trained, unbiased scene graph generator~\cite{knyazev2020graph} in a more practical approach; (ii)~we utilize un-normalized contrastive learning on the SG representation, along with augmentation, to eliminate any potential spurious correlations from annotated data and to heighten the visual information; and (iii)~the use of a GAT encoder to enhance high-level semantic and spatial reasoning on the SG\@.
We further investigate the behavior of visual enhancement when employing a more expressive language encoder, specifically BERT~\cite{kim2018bilinear}.
Importantly, our SelfGraphVQA framework does not require the costly pre-training strategy common to transformer-based models commonly used in vision-language tasks~\cite{wang2022sgeitl,tan2019lxmert,chen2020uniter}.

\section{Related Work}

\begin{table}[tb]
\sisetup{
    round-pad=true,
    round-mode=places,
    round-precision=1,
    table-format=2.1,
}
\centering
\caption{Our experiments revealed a notable accuracy reduction in top-notch methods on the GQA dataset when transitioning from well-annotated to extracted scene graphs. We categorize methods by data type (e.g., annotated data or purely image-question extraction) and SGG usage. All methods are trained and validated uniformly, except for the test extracted configuration, trained on ideal data and validated on extracted SGG data.}
\label{tab:previous}
\footnotesize
\resizebox{\linewidth}{!}{
\begin{tabular}{llS}
    \toprule
    Method & Eval.\ Data & {Acc (\%)} \\
    \midrule
    Human~\cite{gva2019dataset} & {--} & 89.30 \\
    GraphVQA~\cite{liang2021graphvqa} & Annotated/SGG & 94.78 \\
    LRTA~\cite{liang2020lrta} & Annotated/SGG & 93.10 \\
    Lightweight~\cite{nuthalapati2021lightweight} & Annotated/SGG & 77.87 \\
    CRF~\cite{nguyen2022coarse} &  Annotated & 72.10 \\
    LXMERT~\cite{tan2019lxmert} &  Extracted & 59.80 \\
    \midrule
    GraphVQA (original pre-trained on ideal) & \textbf{Test Extracted/SGG} & 29.7 \\
    \midrule
    SelfGraphVQA (Local)  & Extracted/SGG & 51.5 \\
    SelfGraphVQA (Global) & Extracted/SGG & 52.3 \\
    SelfGraphVQA (SelfSim) & Extracted/SGG & 54.0 \\
    \bottomrule
\end{tabular}}
\end{table}

\paragraph{Scene Graph and Visual Question Answering.}
Accurately assessment of VQA tasks, requiring a comprehensive understanding of visual perception and semantic reasoning, has gained substantial attention in the academic community, as these tasks holds significant practical value, particularly in enhancing accessibility for the visually impaired~\cite{ben2017mutan, kim2018bilinear,li2019visualbert,zhang2021vinvl,zellers2022merlot}.

Several works have explored the information that SG representations may bring to VQA~\cite{wang2022vqa,liang2021graphvqa}, as opposed to the more data-hungry transformer-based visual language models~\cite{tan2019lxmert,chen2020uniter,li2019visualbert}.
However, existing SG-VQA approaches typically rely on idealized scene graphs and inherent dataset reasoning~\cite{liang2020lrta,liang2021graphvqa}.
Obtaining such annotations can be costly without an end-to-end pipeline.
Moreover, even SoTA methods in SG-VQA exhibit limited generalization capabilities, potentially due to spurious correlations~\cite{agrawal2023reassessing}.

\paragraph{Self-Supervised Learning.}
Broadly speaking, recent advancements in self-supervised learning can be categorized into normalized~\cite{aitchison2021infonce,chen2020improved} and maximization representation learning~\cite{grill2020bootstrap,chen2021exploring,thakoor2021bootstrapped}.
Contrastive methods aim to bring embeddings of identically labelled images closer together while separating embeddings generated from different versions.
In visual-language data, the prevailing approach for self-supervised learning involves pretraining a transformer-based model on a large dataset to solve pretext tasks before fine-tuning for downstream tasks~\cite{wang2022sgeitl,chen2020uniter,tan2019lxmert,radford2021learning}.
However, these methods can be computationally expensive and complex due to the use of negative samples and masking techniques.
Modern un-normalized contrastive learning methods, \eg, BYOL~\cite{grill2020bootstrap} and SimSiam~\cite{chen2021exploring}, use architectures inspired by reinforcement learning to maximize the informational content of the representations.
In our proposal, we adopt a similarity maximization approach using a Siamese architecture for visual scene graph representation.

\section{Methodology}
We refer the reader to the appendix for the implementation details.
We experiment with the maximization strategy with three independent and distinct similarity losses over either a localized node representation (\ie, object-wise), a global pooled graph representation (i.e., scene-wise), or a regularization node representation term to induce permutation equivariance.
We denote the graph representations $z_i = f_g\big(g(x_i), f_q(q)\big)$, and the predictor's output vectors $p_i = h(z_i)$.
Generally, the representations are maximized by minimizing the generic cosine distance $D$ loss.

\paragraph{Local Similarity.}
\label{meth:local}
To account for permutation invariance in the node representations, we compute cosine distances over all object pairs from the two views and use the maximally similar node embedding pairs to compute the local loss by
\begin{equation}
   L^*_{\mathrm{\ell}}(p_1, z_2) = \frac{1}{O}\sum_{i}^{O} \argmin_{z_{2,j}} D(p_{1,i}, z_{2,j}),
\end{equation}
where $O$ is the number of objects in the scene.
Symmetrically, we compute $L^*_{\mathrm{\ell}}(p_2, z_1)$, to obtain the overall local loss
\begin{equation}
   L_{\mathrm{\ell}}(z_1, z_2) = \frac{1}{2}\big(L^*_{\mathrm{\ell}}(p_1, z_2) + L^*_{\mathrm{\ell}}(p_2, z_1)\big).
\end{equation}

\paragraph{Global Similarity.}
\label{meth:global}
After obtaining a graph representation, we follow an approach similar to cosine similarity maximization for image classification~\cite{grill2020bootstrap,chen2021exploring}.
Along with the intuition that contrasting between global representations may enhance the visual cues, we assume that the global representation contains the full information about the scene.
Similar to the local representation, we minimize the cosine distance, yielding a loss on the form
\begin{equation}
   L_{\mathcal{g}}(z_1, z_2) = \frac{1}{2}  \big(D(p_{1}, z_{2}) + D(p_{2}, z_{1})\big).
\end{equation}

\paragraph{Regularization for Permutation Equivariance.}
\label{meth:anchor}
We employ an \emph{anchor}, where the SG of an unmodified image guides the SG of the augmented image, allowing us to obtain a more accurate representation of the original scene. %
Our assumption is that the local similarity loss decreases the global performance, while global similarity provides a contextual representation but loses local details.
This technique aligns similar nodes and encourages regularization, making augmented scene representations closer to the original, thus mitigating permutation invariance in graph representations.

Denote the anchored representation by $z_1$, and the augmented representation by $z_2$.
We determine intra-similarities of the anchors $s_{1,i} = \argmin_{z_{1,j}} D(z_{1,i}, z_{1,j})$ and similarities of augmented views $s_{2,ij} = D(z_{2,i}, z_{2,j})$.
We then compute cross-entropy (CE) between anchors and augmentations
\begin{equation}
J(z_1, z_2) = \mathrm{CE}(s_1, s_2),
\end{equation}
which acts as a regularizer to constrain permutation equivariance for the augmentations in addition to the local loss. We combine these losses using
\begin{equation}
    L_{\mathcal{s}}(z_1, z_2) = L_{\ell}(z_1, z_2) + J(z_1, z_2),
\end{equation}
which we refer to as a local self-similarity loss (SelfSim).

\paragraph{Distribution Link Representation Regularization.}
Similarly to the regularization for permutation equivariance, we apply link regularization \emph{in conjunction with one of the other three similarity strategies}.
The edges of the \emph{anchor} SG guide the edges of the augmented SG.
Denote the anchored edge score representation by $r_1$, and the augmented edge score representation by $r_2$.
These scores characterize the relationship between the objects in the scene, and we aim to make the link distribution more robust to perturbation.
\emph{In this case, the scene graph generator~\cite{knyazev2020graph} is trainable.}
We compute the cross-entropy between the anchored edge scores and the augmented edge scores $J_e(r_1, r_2) = \mathrm{CE}(r_1, r_2)$, which acts as a regularizer to constrain the link prediction distribution, yielding
\begin{equation}
    L_{\mathcal{e}}(z_1, z_2) = L_{\ell}(z_1, z_2) + J_e(r_1, r_2).
\end{equation}
All models utilizing this added link distribution regularizer are characterized by the inclusion of the term ``link.''

\paragraph{Overall Optimization Objective.}
Lastly, we outline the overall loss for optimizing the VQA objective.
To identify the correct answer $a \in A$ given an example $(x, q, A)$, where $x$ represents the input image, and $q$ is the associated question, we extract a point estimate of probabilities
\begin{equation}
    p(a \given x, q) = \sigma \left( \logit(x) \right),
\end{equation}
where $\sigma$ is the softmax function, and $\logit(x) = f(x,q)$ are the logits for all possible answers produced by our encoder.
We calculate the cross-entropy loss for each instance,
\begin{equation}
    L_{\mathcal{sup}}(x) = \mathrm{CE} \left( p(a \given x, q), a \right).
\end{equation}
Our combined training loss is then given by
\begin{equation}
    L(x) = \alpha L_{\mathcal{sup}}(x) + \beta L'(z_1, z_2),
\end{equation}
where $L'$ can be any of the aforementioned similarity loss strategies: $L_\ell$, $L_\mathcal{g}$, or $L_\mathcal{s}$, with or without $L_\mathcal{e}$.
The $\alpha$ and $\beta$ are controlled hyperparameters that balance the contribution of the various components in the total loss.

\section{Experiments and Ablations}
We evaluate our framework on the GQA dataset~\cite{gva2019dataset}.
Our study aims to establish a practical foundation for demonstrating the potential of SG along with an un-normalized contrasting approach to improve visual cues for VQA\@.
Despite the noise data in the extracted SG, we demonstrate its effectiveness, Fig.~\ref{tab:sota}, by highlighting the importance of further exploration.
The utilization of non-idealized SG-VQA methods with un-normalized contrastive learning leads to improvements across all metrics, Table~\ref{tab:sota}.
Furthermore, our framework demonstrates faster convergence during training, approximately 20\% faster in epochs compared to baselines.
However, further investigation is required to validate them.

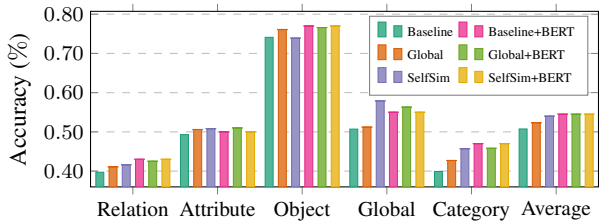
\begin{figure}[tb]
\centering
\scriptsize
\begin{tikzpicture}
\begin{axis}[
  footnotesize,
  width=\linewidth,
  height=0.47\linewidth,
  yticklabel style={%
    /pgf/number format/fixed,
    /pgf/number format/fixed zerofill,
    /pgf/number format/precision=2,
  },
  /pgfplots/error/.style={
    error bars/.cd,
    y dir=both, y explicit relative,
  },
  ybar,
  /pgf/bar width=3pt,
  xtick={0,1,...,6},
  xticklabels={Relation, Attribute, Object, Global, Category, Average},
  ylabel={Accuracy ($\%$)},
  enlarge x limits={abs=.5},
  dark list,
  legend style={
    legend columns=3,
    transpose legend,
    draw=gray,
    font=\tiny,
  },
  legend cell align={left},
  legend style={at={(0.98,0.975)},anchor=north east},
  legend entries = {Baseline, Global, SelfSim, Baseline+BERT,Global+BERT, SelfSim+BERT},
]
\addplot+[fill] table[y=Baseline, x index=0, y error=Baseline_std, col sep=comma] {images/struct-types.csv};
\addplot+[fill, /pgfplots/error] table[y=Global, x index=0, y error=Global_std, col sep=comma] {images/struct-types.csv};
\addplot+[fill, /pgfplots/error] table[y=SelfSim, x index=0, y error=Global_std, col sep=comma] {images/struct-types.csv};
\addplot+[fill, /pgfplots/error] table[y=SelfSim+BERT, x index=0, y error=Baseline+BERT_std, col sep=comma] {images/struct-types.csv};
\addplot+[fill, /pgfplots/error] table[y=Global+BERT, x index=0, y error=Baseline+BERT_std, col sep=comma] {images/struct-types.csv};
\addplot+[fill, /pgfplots/error] table[y=SelfSim+BERT, x index=0, y error=Baseline+BERT_std, col sep=comma] {images/struct-types.csv};

\end{axis}
\end{tikzpicture}%
\caption{Accuracy on different question types.}
\label{fig:breakdown}
\end{figure}

The un-normalized contrastive approach universally enhances results across question categories (Fig.~\ref{fig:breakdown}), with specific types of approaches further improving the model's performance based on the query type.

\begin{table}[tb]%
\sisetup{
    round-pad=true,
    round-mode=places,
    round-precision=1,
    table-format=2.1,
}
\centering
\scriptsize
\caption{Results (\%) on GQA by standard metrics.}
\resizebox{\linewidth}{!}{%
\begin{tabular}{lS@{ }@{}S@{ }@{}S@{ }@{}S@{ }@{}S@{ }@{}S@{ }@{}S}
    \toprule
    Method & {Binary ($\uparrow$)} & {Open ($\uparrow$)} & {Consist. ($\uparrow$)} & {Validity ($\uparrow$)} & {Plausab. ($\uparrow$)} &  {Distr. ($\downarrow$)} & {Acc ($\uparrow$)}\\
    \midrule
    Baseline & 65.8 & 29.7 & 58.2 & 94.9 & 90.5 & 11.7 & 50.1\\
    Baseline+BERT & 68.0 & 32.2 & 62.56 & 95.0 & 90.9 & 7.7 & 53.8\\
    \midrule
    Local & 66.80 & 30.15 & 59.42 & 94.88 & 90.61 & 8.83 & 51.47\\
    Global & 67.74 & 30.76 & 62.46 & 94.9 & 90.6 & 6.74 & 52.33\\
    SelfSim & \textbf{68.4} & 31.3 & \textbf{65.9} & 94.9 & 90.7 & \textbf{2.1} & 54.0\\
    \midrule
    Global+BERT+link & 68.0 & \textbf{33.0} & 63.9 & 95.0 & \textbf{91.2} & 8.9 & \textbf{54.5}\\
    SelfSim+BERT+link & 68.2 & 32.8 & 64.3 & \textbf{95.0} & 91.0 & 8.0 & \textbf{54.5}\\
    \bottomrule
    \label{tab:sota}
\end{tabular}
}%
\vspace*{-15pt}
\end{table}

We conducted ablations to demonstrate the functionality of our approach and carried out detailed observations that go beyond mere reliance on metrics using the GQA dataset.

\paragraph{Does the Scene Graph Really Matter?}
Through a perturbation study where images were augmented based on question types, we introduced disruptive noise such as image flipping to challenge the model's ability to answer spatial relational questions. The goal was to observe mistakes in the model's answers. The results, compared to the baseline (Table~\ref{tab:resultsequestion}), showed greater variation in our model's performance, indicating that it pays more attention to visual information, whereas the baseline appears to rely on other sources of information.

\begin{table}[tb]
\sisetup{
    round-pad=true,
    round-mode=places,
    round-precision=1,
    table-format=2.1,
}
\centering
\caption{Change in accuracy under potentially disruptive augmentations and perturbations.}
\footnotesize
\resizebox{\linewidth}{!}{%
\begin{tabular}{lrSSSS}
    \toprule
    Question Type & Augmentation & {Baseline} & {Global} & {Local} & {SelfSim}\\
    \midrule
    Relation & Flip & -1.6 & -3.37 & -3.24 & -3.89\\
    Attribute & Strong Color Jitter &  {+1.14} & -3.72 & -0.76 & -1.15\\
    Global & Gaussian Noise + Crop & -5.6 & -7.7 & -5.45 & -8.1\\
    \bottomrule
\end{tabular}}
\label{tab:resultsequestion}
\end{table}

\paragraph{Are Performance Gains Mainly Due to Augmentations?}
We compared our approach with the baseline architecture, training solely with data augmentation techniques to evaluate their influence on overall performance. Table~\ref{fig:selfxaug} provides evidence that data augmentation techniques actually impair the performance of the architecture.

\begin{table}[tb]
\sisetup{
    round-pad=true,
    round-mode=places,
    round-precision=1,
    table-format=2.1,
}
    \footnotesize
    \centering
    \caption{Results(\%) of the Aug. Baseline and SelfSim.}
    \resizebox{.8\linewidth}{!}{
    \begin{tabular}{lSSSSS}
        \toprule
        Method & {Binary} & {Open} & {Validity} & {Plausibility} & {Acc}\\
        \midrule
        Baseline Aug & 65.05 & 28.72 & 94.62 & 90.08 & 50.11\\
        SelfSim & 68.44 & 31.28 & 94.9 & 90.7 & 54.0 \\
        \bottomrule
    \end{tabular}
    }
    \label{fig:selfxaug}
    \vspace*{-10pt}
\end{table}

\paragraph{Are Our Models Less Biased?}
Our initial hypothesis was that current top-performing models might incorporate biases present in the questions into their weights.
We conducted experiments to analyze this issue, introducing random noise to features in the scene graph while preserving its topology, and perturbing the language in up to 50\% of the words in the questions.
The results in Table~\ref{tab:bias} demonstrate that our approach relies less on linguistic features, prioritizing overall information and reducing linguistic bias. Additionally, we explored visual enhancement, even when trained with a more expressive language module such as BERT. The experiments in Table~\ref{tab:bias} examine the impact of using BERT and its effect on enhancing visual information.

\begin{table}[tb]
\sisetup{
    round-pad=true,
    round-mode=places,
    round-precision=1,
    table-format=2.1,
}
\centering
\caption{Sensitivity of accuracy (\%) for bias question analyzes of SelfGraphVQA and SelfGraphVQABERT.}
\label{tab:bias}
\footnotesize
\resizebox{\linewidth}{!}{%
\begin{tabular}{lSSSS}
    \toprule
    \textbf{Setup} & & \textbf{Methods} & \\
    \midrule
    Scene Graph + Question  & {Baseline} & {Local} & {Global} & {SelfSim}\\
    \midrule
    Noise + SG  & 16.2 & 16.60 & 28.6 & 26.6\\
    Question + Noise  & 39.93 & 38.304 & 37.39 & 39.79\\
    Noise + Noise & 12.7 & 14.6 & 18.9 & 21.0\\
    \toprule
    Question + Scene Graph  & {BERT Baseline} & {BERTGlobal+link} & {BERTSelfSim+link}\\
    \midrule
    Noise + SG & 21.0 & 23.2 & 24.5\\
    Question + Noise  & 42.4 & 41.8 & 42.8\\
    Noise + Noise & 19.8 & 21.7 & 21.3\\
    \bottomrule
\end{tabular}
}
\vspace*{-10pt}
\end{table}

\begin{figure}[!tb]
    \newlength{\sz}
    \setlength{\sz}{.33\linewidth}
    \centering
    \scriptsize
    \resizebox{\linewidth}{!}{%
    \begin{tabular}{%
    lll
    }
        Relative & Synonym & Ambiguous \\
        \includegraphics[width=\sz]{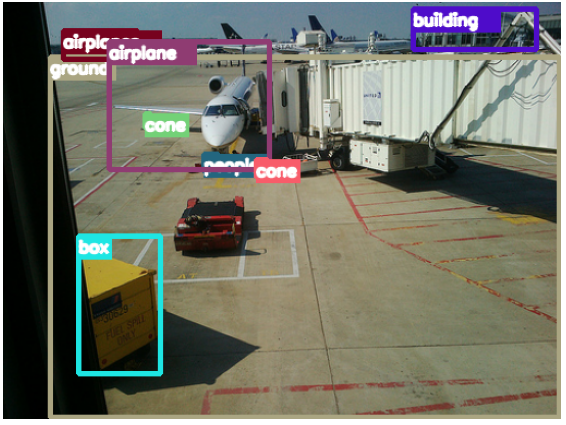} &
        \includegraphics[width=\sz]{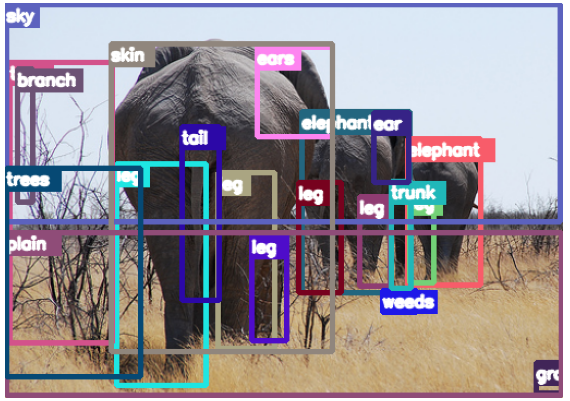} &
        \includegraphics[width=\sz]{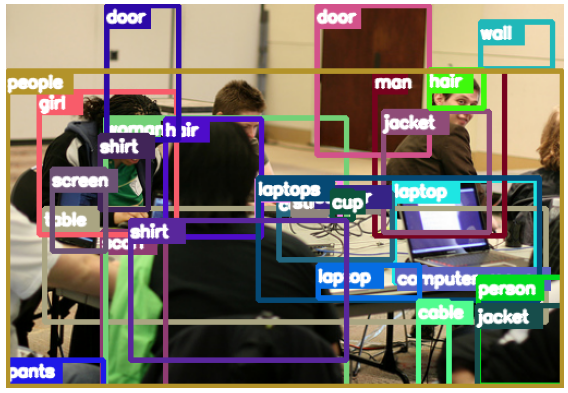} \\
        \includegraphics[width=\sz]{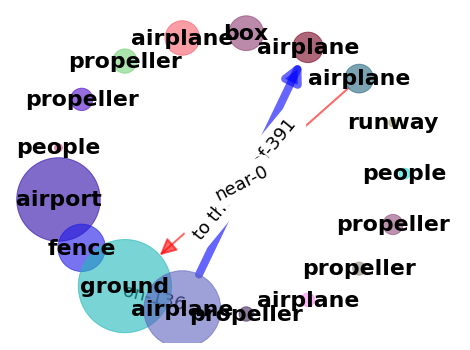} &
        \includegraphics[width=\sz]{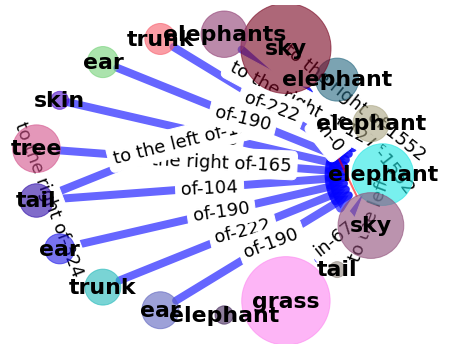} &
        \includegraphics[width=\sz]{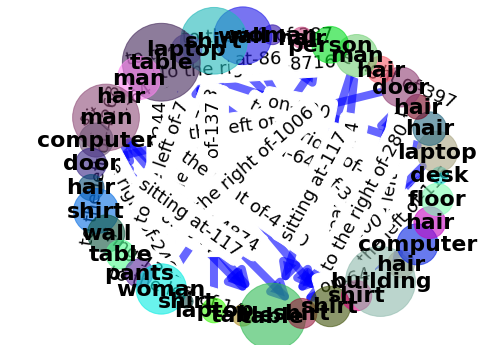} \\
        Q: Is there an airplane in the &
        Q: Where are the weeds? &
        Q: Is the man to the right \\
        picture that is not small? &
        &
        or to the left of the cup?
        \\
        Answer: Yes &
        Answer: Plain&
        Answer: Right  \\
        Prediction: No &
        Prediction: Field &
        Prediction: Left \\
    \end{tabular}}
    \caption{Examples to demonstrate the complexity of VQA.}
    \vspace*{-10pt}
    \label{fig:ablation}
\end{figure}

\paragraph{Examples.}
Given the wide range of acceptable answers, we argue that solely relying on standard evaluation metrics may not provide a fair comparison, thus presenting additional challenges to the field.
Fig.~\ref{fig:ablation} demonstrates the utility of SG for interpretability, as they enable a graphical analysis of objects and the overall composition of the scene.

\section{Conclusions}
\label{sec:conclusions}
Despite promising results in VQA tasks with idealized SG, our study revealed that models relying on manually annotated and expensive SG struggle with real-world data.
To address this, we proposed SelfGraphVQA, a more practical SG-VQA framework that breaks the spurious correlation of annotated SG and learns to answer questions using extracted SG from a pre-trained SG generator.
We employed un-normalized contrastive learning to maximize similar graph representations in different views.
All approaches utilizing self-supervision showed improvement over their baselines.
Overall, we demonstrated the effectiveness of extracted SG in VQA, underscoring the significance of continued exploration of the potential of SG for complex tasks.
We also showed that self-supervision over the SG representation improved the results by enhancing the visual information within the task.
We hope that this work raises awareness of the challenges of accentuating the role of the scene in answering questions from images.

\section*{Acknowledgements}

This work was supported in part by the FAPESP (São Paulo Research Foundation) grant no.~2022/09849-8 and BEPE grant no.~2022/09849-8.
The computations were performed in part on resources provided by Sigma2---the National Infrastructure for High-Performance Computing and Data Storage in Norway---through Project~NN8104K.
This work was funded in part by the Research Council of Norway, via the Centre for Research-based Innovation funding scheme (grant no.~309439), and Consortium Partners.

{\small
\bibliographystyle{ieee_fullname}
\bibliography{abrv,egbib}
}

\input{appendix-content}

\end{document}

%% file: appendix-content.tex
\appendix
\counterwithin{figure}{section}
\counterwithin{table}{section}

\section{Datasets}

We evaluate our SelfGraphVQA frameworks on the GQA dataset~\cite{gva2019dataset}.
GQA is another large-scale effort (22M questions, each with
one answer) that focuses on the compositionality of template-generated questions for real-world images.
We use the official train/validation split of GQA.

The GQA dataset was selected for evaluation because it includes complex relational and spatial questions that require multiple reasoning skills, spatial understanding, and multi-step inference.
These characteristics make it more challenging compared to previous visual question-answering datasets.
Consequently, the GQA dataset is well-suited for evaluating the performance of scene graph models.

In contrast to prior studies~\cite{nguyen2022coarse,agrawal2023reassessing}, our approach takes a simplistic approach by considering solely the ground truth distribution as a potential answer in the training dataset as a candidate for the answer distribution, without any filter techniques.
Table~\ref{tab:stats} provides detailed statistics for each dataset examined in our investigation.

\begin{table}[tb]
  \sisetup{
    round-pad=true,
    round-mode=places,
    round-precision=1,
    table-format=2.1,
  }
  \centering
  \caption{Detailed statistics for the GQA dataset examined in our study compared to other possible statistics and the original paper dataset.}
  \label{tab:stats}
  \footnotesize
    \begin{tabular}{{lr}}
      \toprule
      & Answers Candidates \\
      \midrule
      Ours & 1878\\ Alternative~\cite{agrawal2023reassessing,lu2019vilbert,nguyen2022coarse}& 1533\\
      Original~\cite{gva2019dataset,goyal2017making} & 1878\\
      \bottomrule
    \end{tabular}
  \end{table}

  Despite the substantial variations in the answering classes, we emphasize that our method proves to be effective and comparable to other existing approaches.
  In addition to the aforementioned points, this further highlights the fact that VQA is a complex and expansive challenge that lends itself to various approaches and needs continued exploration and refinement.

  \section{Baseline Architecture}

\begin{figure}
    \centering
      \begin{tikzpicture}[
        node distance=.5cm,
        proc/.style={
          draw,
          fill=#1,
          rectangle,
          rounded corners,
          text width=2.5cm,
          minimum height=1.0cm,
          align=center,
          font=\footnotesize,
        },
        proc/.default=white,
        edge/.style={
          shorten <=2pt,
          shorten >=2pt,
          ->,
          rounded corners,
        },
        ]
        \node (image) [rectangle, green!25, text=black] {\includegraphics[width=2cm]{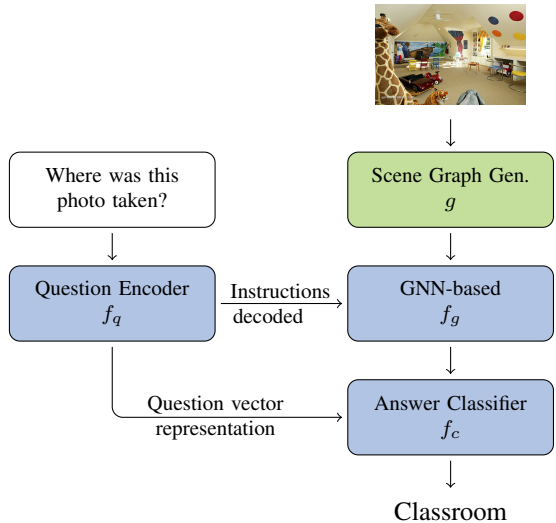}};
        \node (graph) [proc=LimeGreen!50, below=of image] {Scene Graph Gen.\\$g$};
        \node (question) [proc, left=1.75cm of graph] {Where was this \\ photo taken?};
        \node (question-encoder) [proc=NavyBlue!30, below= of question] {Question Encoder\\$f_q$};
        \node (ggnn) [proc=NavyBlue!30, below= of graph] {GNN-based \\$f_g$};
        \node (classifier) [proc=NavyBlue!30, below= of ggnn] {Answer Classifier \\$f_c$};
        \node (answer) [text width=2cm, below= of classifier, align=center] {Classroom};

        \draw [edge] (image) -- (graph);
        \draw [edge] (question) -- (question-encoder);
        \draw [edge] (graph) -- (ggnn);
        \draw [edge] (question-encoder) --  (ggnn) node[pos=.4, font=\footnotesize, text width=1cm, align=center] {Instructions\\decoded};

        \draw [edge] (question-encoder.south) |- (classifier) node[midway, font=\footnotesize, text width=2.5cm, align=center, anchor=west] {Question vector\\representation};
        \draw [edge] (ggnn) -- (classifier);
        \draw [edge] (classifier) -- (answer);
      \end{tikzpicture}

      \caption{The baseline architecture.}
      \label{fig:architecture}
    \end{figure}

  Figure~\ref{fig:architecture} depicts the overall components of our baseline architecture.
  The unbiased pre-trained scene graph generator model utilized in this study originates from the research paper authored by Knyazev et al.~\cite{knyazev2020graph}.
  Applying a density-normalized edge loss to the model, the authors contend that the model is aware of the graph density and, therefore, generalizes better even to rare compositions.

  In our project, the frozen weights pre-trained Scene Graph Generator takes the image information and generates a scene graph representation.
  The Question Encoder receives the instructions and provides them to the GNN-based encoder.
  Each layer of the module pays attention to these instructions in order to update its hidden node states.
  The Classifier then takes the graph representation and the question vector concatenates them, and predicts the correct answer.

    We use the similar architecture of the state-of-the-art graph-based GraphVQA model~\cite{liang2021graphvqa} and LRTA~\cite{liang2020lrta} over the GQA dataset as a baseline for our experiments, with some modifications in order to reduce the dependence on the annotated available data, as we aim to mitigate the limitations imposed by data availability and enhance the model's generalizability.

    For practical purposes, the functional program instructions accompanying each question in the GQA dataset~\cite{gva2019dataset} are not necessarily available for inference on real-world data, so we train our decoder to decode the instructions from the question itself.
    These additional labels are processed by the reasoning module in the GraphVQA model which we explicitly omit in our baseline, as we are more interested in generalizability and real-world performance rather than expressively \emph{solving} the GQA dataset.

    In addition, we omit the pre-processing using the scene graph encoding module of the original GraphVQA, as the scene graph generation model $g$ was selected to extract high quality SG-representations.
    Here, our $f_g$ module is a graph attention network (\eg GAT)~\cite{velivckovic2017graph}.

    In the GloVE embedding design, both the query encoder $f_q$ and the graph encoder $f_g$ designs are shared between the original baseline and our proposed modified model.
    Whereas in the BERT design, we only take the similarity of the graph encoder module $f_g$ design, as our query encoder $f_q$ and the language embedding is a BERT model.
    By adapting the similar SoTA architecture strategy to the specific design choices of each model, we aim to evaluate the performance and effectiveness of our proposed approach.

    \section{Architecture Details}

    Within this section, we aim to provide additional details regarding all components of our implementation approaches.

    To ensure clarity and facilitate better comprehension, we have divided this section into two subsections: one discussing the utilization of GloVE word embedding along with a transformer-based model for the question encoder, and the other focusing on the application of BERT for word embedding and the question encoder.

    Table~\ref{tab:compared} provides a comprehensive overview of the two approaches employed in this study.

    It is worth mentioning that the scene graph generator module has its weights frozen in all training approaches, except when we employ the Distribution Link Representation Regularization technique.

    \subsection{GloVE Word Embedding and Transformer-based Question Encoder}
    The images are fed through a pre-trained scene graph generator $g$ from~\cite{knyazev2020graph} work that generates scene graphs from images on the fly.

    Except for the pooled graph-level representation (\ie, the module that feeds the classifier), which has a dimension size of \num[drop-zero-decimal=true]{512}, all node and edge features have dimension size \num[drop-zero-decimal=true]{300}.

    The word embedding for the transformed-based query encoder module $f_q$ has its initial weights initialized by using embeddings from GloVe~\cite{pennington2014glove}.
    Both hidden states and word embedding vectors have a dimension size of \num[drop-zero-decimal=true]{300}.
    The question representation is produced by the transformed-based question encoder.

    Following \cite{liang2020lrta, liang2021graphvqa} work, we adopt a hierarchical sequence generation design, \ie, a transformer decoder model first parses the question into a sequence of $M$ instruction vectors, $[i_1, i_2, \dots, i_M]$.
    The $i$-th instruction vector will correspond exactly to the $i$-th execution step processed by the GNN encoder $f_g$ module.
    In our experiments, we force $M$ equals five.
    We note that SelfGraphVQA does not require any explicit supervision on how to solve the instruction step from the question, and we only supervise the final answer prediction.

    For the un-normalized contrastive approach, the MLP prediction head $h$ plays a crucial role in our model architecture.
    It comprises three fully connected layers, each followed by batch normalization and ReLU activation, except for the final layer.
    This setup ensures non-linearity and facilitates effective feature extraction.
    It is important to note that the MLP prediction head is exclusively utilized during the training phase and is subsequently discarded during inference, which aligns with prevailing practices in contemporary self-supervised training methods~\cite{chen2021exploring,grill2020bootstrap}.

    The classification module $f_c$ is another integral component of our model.
    It is designed as a two-layer MLP with a dropout rate of $0.2$ and ELU activation.

    As explained in Section~3, we independently apply the three self-supervised losses (\ie, local similarity, global similarity, and regularization for permutation equivariance) and compared performances.
    Our experimental choices were designed to minimize possible biases in the evaluation of our proposed framework.

    Both anchored and augmented scene graphs along with the question ground on the scene feed our encoder model to infer a predicted answer.
    For a fair comparison, we train most of our model from scratch, except for the pre-trained scene graph generator $g$, whose weights are frozen.

    \subsection{BERT Word Embedding and Question Encoder}

    In this case, we employ the BERT model as our word embedding approach and the question encoder, as being a more expressive language model.

    Once again, the images are fed through a pre-trained scene graph generator $g$ from \cite{knyazev2020graph} work that generates scene graphs from images on the fly.
    In this particular case, all graph-level and node-level representations possess a dimension size of \num[drop-zero-decimal=true]{512}, encompassing both node and edge features.
    This configuration is deliberately chosen to ensure that the dimensions of the representations closely align with the dimension yielded by BERT word embedding, which is \num[drop-zero-decimal=true]{756}.
    By maintaining consistency in the dimensionality across different components, we aim to facilitate seamless integration and compatibility with BERT-based models.

    \begin{table}[tb]
      \sisetup{
        round-pad=true,
        round-mode=places,
        round-precision=0,
        table-format=2,
      }
      \centering
      \caption{Detailed dimensions used in our study when employing the GloVE and BERT approaches.}
      \label{tab:compared}
      \footnotesize
      \resizebox{\linewidth}{!}{%
        \begin{tabular}{lSSSSS}
          \toprule
          Methods & {Word dim.} & {Question dim} & {Node Dim} & {Link Dim} & {Graph dim}\\
          \midrule
          GloVE+Transf & 300 & 300 & 300 & 300 & 512\\
          BERT & 756 & 512 & 512 & 512 & 512\\
          \bottomrule
        \end{tabular}
      }
    \end{table}

    The word embedding for the BERT query encoder $f_q$ has its initial weights initialized by using embeddings from BERT~\cite{kim2018bilinear}.
    Both hidden states and word embedding vectors have a dimension size of \num[drop-zero-decimal=true]{512}.
    The final question representation is derived by taking the average of all word embedding representations generated by BERT.

    Following the same approach of \cite{liang2020lrta,liang2021graphvqa}, we adopt a hierarchical sequence generation design, \ie, a transformer decoder module first parses the encoded question into a sequence of $M$ instruction vectors, $[i_1, i_2, \dots, i_M]$.
    The $i$-th instruction vector will correspond exactly to the $i$-th execution step processed by the GNN encoder $f_g$ module.
    In our experiments, we force $M$ equals five.
    We note that SelfGraphVQA does not require any explicit supervision on how to solve the instruction step from the question, and we only supervise the final answer prediction.

    In this scenario, we employ two self-supervised loss techniques: global similarity and regularization for permutation equivariance.
    Additionally, we incorporate the Distribution Link Representation Regularization method overall approaches performed in this case.
    It is important to note that the Distribution Link Representation Regularization is jointly executed with one of the self-supervised loss techniques.

    As mentioned earlier, in this case, except for the object detector within the module, we have unfrozen the scene graph generator $g$ weights, allowing it to be trainable and to learn the representation and classification during the training process, merely according to the prediction answers.
    We have made deliberate experimental choices to mitigate potential biases and ensure an unbiased evaluation of our proposed framework.

    For the un-normalized contrastive training step, we employ the MLP prediction head $h$.
    It comprises three fully connected layers, each followed by batch normalization and ReLU activation, except for the final layer.
    This setup ensures non-linearity and facilitates effective feature extraction.
    It is important to note that the MLP prediction head is exclusively utilized during the training phase and is subsequently discarded during inference, which aligns with prevailing practices in contemporary self-supervised training methods~\cite{chen2021exploring,grill2020bootstrap}.

    The classification module $f_c$ is another integral component of our model.
    It is designed as a two-layer MLP with a dropout rate of $0.2$ and ELU activation.

    \section{Training Details}
    In this section, we provide further elaboration on our training approaches.
    Likewise, we have divided this section into two subsections: one with the utilization of GloVE word embedding along with a transformer-based model for the question encoder, and the other focusing on the application of BERT for word embedding and the question encoder.

    \begin{table}[tb]
      \sisetup{
        round-pad=true,
        round-mode=places,
        round-precision=0,
        table-format=2,
      }
      \centering
      \caption{Training details for the GloVE and BERT approaches employed in our study.}
      \label{tab:trainingdetails}
      \footnotesize
      \begin{tabular}{lrrrr}
        \toprule
        Methods & {Batch} & Optimizer & {lr} & {Epochs}\\
        \midrule
        GloVE+Transf & 64 & Adam & $10^{-4}$ & 50\\
        BERT & 32 & Adam Belief & $10^{-4}$ & 50\\
        \bottomrule
      \end{tabular}
    \end{table}

    \subsection{GloVe Word Embedding and Transformer-based Question Encoder}

    We train the models using the Adam optimizer with a learning rate of $10^{-4}$ and weight decay $10^{-4}$. We apply a batch size of \num[drop-zero-decimal=true]{64}, and a a linear learning rate schedule using a factor of $10^{-1}$ for every \num[drop-zero-decimal=true]{20} epochs.
    All models are trained for \num[drop-zero-decimal=true]{50} epochs.
    We emphasize that during training the weights of the scene graph generator $g$ are frozen, and do not receive weight updates.

    \subsection{BERT for Word Embedding and Question Encoder}

    We train the models using the Belief Adam optimizer with a learning rate of $10^{-4}$ and weight decay $10^{-4}$.
    We apply a batch size of \num[drop-zero-decimal=true]{32}, and a linear learning rate schedule using a factor of $10^{-1}$ for every \num[drop-zero-decimal=true]{10} epochs.
    All models are trained for \num[drop-zero-decimal=true]{50} epochs.
    It is worth noting that in these cases, the weights of the scene graph generator $g$ are not frozen during training.
    This deliberate choice allows for continual updates and improvements, particularly in the edge representation, through the utilization of the Distribution Link Representation Regularization strategy.

    \section{Self-Supervised implementation details}
    Table~\ref{tab:selfimplementation} provides a comprehensive overview of the approach adopted in our study.
    It is worth noting that our training process was conducted sequentially and iteratively, allowing us to evaluate the performance of each approach before deciding on the subsequent implementation choice.

    For instance, upon observing that the Local Similarity approach exhibited comparatively lower performance, albeit surpassing the baseline, we made the decision to discontinue its implementation on further research (i.e. with the BERT module and link distribution regularization approach).
    This strategy narrowed down the training possibilities, enabling us to focus solely on the most promising experiments.
    Another noteworthy example pertains to the utilization of BERT as our word embedding and query encoder module.
    Upon observing its positive impact on results, we exclusively applied the link distribution regularization technique with this architecture.

    \begin{table}[tb]
      \sisetup{
        round-pad=true,
        round-mode=places,
        round-precision=0,
        table-format=2,
      }
      \centering
      \caption{Detailed self-supervised implementation in our study by approaches.}
      \label{tab:selfimplementation}
      \footnotesize
      \resizebox{\linewidth}{!}{%
        \begin{tabular}{llcccc}
          \toprule
          & SGG Methods & Baseline & Local Sim & Global Sim. & Self Sim \\
          \midrule
          \multirow{2}{*}{GloVE+Transf}  & Frozen SGG & \checkmark & \checkmark & \checkmark  & \checkmark  \\
          & Link Regularizer &  &   &   &  \\
          \multirow{2}{*}{BERT} & Frozen SGG & \checkmark &  &  \checkmark &  \\
          & Link Regularizer &  &   & \checkmark & \checkmark \\
          \bottomrule
        \end{tabular}
      }
    \end{table}

    \section{Further Ablations}

    \subsection{Further Discussion on Language Bias}
    We elaborate on additional experiments aimed at evaluating the model's robustness when trained with the BERT module.
    In this case, the experiments investigate the impact of using a more expressive language model, such as BERT, on language biases in the VQA task and whether it harms the enhancement of visual information.
    We evaluate both how the biases convey not ideal information when using a more expressive language model such as BERT, and how the self-supervised approaches perform for robustness\@.

    In this particular experiment, we augmented the images using various semantically-preserving techniques including Gaussian blur, Gaussian noise, color jitter with adjustments to brightness, contrast, and hue, as well as random rotation of up to 45 degrees.
    As for the questions, a similar approach was employed by randomly replacing up to 50\% of the words with other words.ù

    In this context, we emphasize that our approach  maintains the semantic integrity of the image content.
    Consequently, the underlying model retains its fundamental objective of accurately predicting the correct response, despite the heightened complexity introduced.

    Table~\ref{tab:bias2} demonstrated that even when employing a more expressive language model in the GQA dataset, the self-supervised learning still enhances the visual information for the predicted answer.
    Precisely, the results presented indicate that our approaches exhibit greater resilience to noise while maintaining the importance of visual information for the task.

    We emphasize that the findings of this study demonstrate that despite the integration of a more expressive language model, such as BERT, the self-supervised learning method remains effective in leveraging visual data to classifier the predicted answers.
    Nevertheless, it is crucial to highlight that in this particular scenario, the results indicate a possible influence of language biases inherent in the dataset when utilizing a more advanced language model.
    See the results when the perturbation is employed solely on the scene graph compared to the non-perturbed one, in Table~\ref{tab:bias2}.
    Additionally, when analyzing the results with full perturbation, the findings indicate an enhanced level of robustness when the self-supervision technique is combined with the model.

    \begin{table}[tb]
      \sisetup{
        round-pad=true,
        round-mode=places,
        round-precision=1,
        table-format=2.1,
      }
      \centering
      \caption{Sensitivity of accuracy (\%) for bias analyzes of BERT module.}
      \label{tab:bias2}
      \footnotesize
      \resizebox{\linewidth}{!}{%
        \begin{tabular}{lSSS}
          \toprule
          \textbf{Setup} & & \textbf{Methods} &  \\
          \midrule
          Question + Scene Graph  & {BERT Baseline} & {BERTGlobal+link} & {BERTSelfSim+link}\\
          \midrule
          Noise + SG & 21.0 & 23.2 & 24.5\\
          Question + Noise  & 42.4 & 41.8 & 42.8\\
          Noise + Noise & 19.8 & 21.7 & 21.3\\
          \bottomrule
        \end{tabular}
      }
    \end{table}

    \subsection{Does SelfGraphVQA have a few-shot learning capability?}

    We trained SelfGraphVQA with varying percentages of labeled data and found comparable performance to the GQA dataset, suggesting that adding self-supervised contrastive loss improves model generalization.
    We wanted to evaluate the different models on subsets of the full dataset.
    We tested reducing the ground truth labeling requirements and compared the performance when using SelfGraphVQA as opposed to directly training a fully supervised classification network.

    In this case, we trained our SelfGraphVQA varying the percentage of labeled data, (\ie, 20\%, 50\%, and 100\% of data) and evaluated it on the test dataset.
    As demonstrated in Fig.~\ref{fig:per-data}, our proposal performs comparably with half of the GQA dataset evaluated on standard metrics. This insinuates that adding self-supervised un-normalized contrastive loss improves the generalization of the model.

    Table~\ref{tab:sotapercentage} shows how our proposal performs with the standard metrics when trained with 50\% of training data, and we see that the three approaches perform on par with the baseline trained on the full dataset. In particular, the validity and plausibility metrics are consistent when compared to models trained on the full dataset.

    Our intuition is that these metrics relate to linguistic bias and do not necessarily require large amounts of samples to converge, indicating that the model learns with little data what type of answer it should guess based on the type of question.

    \pgfmathsetmacro{\tmp}{-1}
    \global\def\ci{\tmp}
    \newcommand{\plotwithshade}[1]{%
      \pgfmathparse{int(\ci+1)}
      \global\def\ci{\pgfmathresult}
      \pgfmathsetmacro{\s}{\ci*5}
      \pgfplotsset{cycle list shift=-\s}
      \addplot+[] table[y=#1, x=per, y error=Baseline_std, col sep=comma] {images/per-data.csv};

      \addplot[name path=#1-upper, draw=none] table[x=per, y expr=\thisrow{#1}+\thisrow{#1_std}, col sep=comma] {images/per-data.csv};
      \addplot[name path=#1-lower,draw=none] table[x=per, y expr=\thisrow{#1}-\thisrow{#1_std}, col sep=comma] {images/per-data.csv};
      \pgfmathsetmacro{\s}{int(3*\ci+3)}
      \pgfplotsset{cycle list shift=-\s}
      \addplot+[fill, opacity=.2] fill between[of=#1-upper and #1-lower];
    }

    \begin{figure}[tb]
      \centering
        \begin{tikzpicture}
          \begin{axis}[
            footnotesize,
            width=\linewidth,
            height=0.65\linewidth,
            xlabel={Percentage of Data},
            ylabel={Accuracy (\%)},
            ymajorgrids=true,
            grid style=dotted,
            xtick={20,30,...,100},
            error setup/.style={
              error bars/.cd,
              y dir=both,
              y explicit,
            },
            dark marker list,
            legend style={
              legend columns=5,
              draw=gray,
              font=\scriptsize,
            },
            legend cell align={left},
            legend style={at={(0.5,1.05)},anchor=south},
            ]

            \addplot+[dashed] table[y=Baseline, x=per, y error=Baseline_std, col sep=comma] {images/per-data.csv};
            \addplot+[error setup] table[y=Local, x=per, y error=Local_std, col sep=comma] {images/per-data.csv};
            \addplot+[error setup] table[y=Global, x=per, y error=Global_std, col sep=comma] {images/per-data.csv};
            \addplot+[error setup] table[y=SelfSim, x=per, y error=SelfSim_std, col sep=comma] {images/per-data.csv};
            \legend{Baseline, Local, Global, SelfSim}
          \end{axis}
        \end{tikzpicture}
      \caption{Evaluation curve by percentage of data used in training on GQA dataset. The models obtain comparable results to baseline with 50\% of the data. Note that we only illustrate the accuracy of the baseline trained on the full dataset for reference purposes.}
      \label{fig:per-data}
    \end{figure}
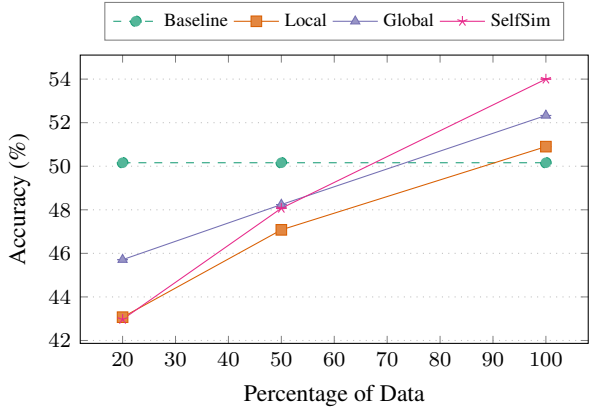

    \begin{table}[tb]
      \sisetup{
        round-pad=true,
        round-mode=places,
        round-precision=1,
        table-format=2.1,
      }
      \centering
      \caption{Results (in \%) evaluating by the standard metrics when training with 50\% of GQA dataset.}
      \label{tab:sotapercentage}
      \footnotesize
      \resizebox{\linewidth}{!}{
        \begin{tabular}{lSSSSSS}
          \toprule
          Method & {Binary} & {Open} & {Consistency} & {Validity} & {Plausibility} & {Accuracy}\\
          \midrule
          Global & 63.52 & 27.58 & 54.14 & 94.83 & 90.10 & 48.24\\
          Local & 63.52 & 25.63 & 51.62 & 94.56 & 89.30 & 47.08\\
          SelfSim & 64.28 & 27.33 & 54.68 & 94.77 & 90.11 & 48.07\\
          \bottomrule
      \end{tabular}}
    \end{table}

    \subsection{More Examples}
\begin{figure*}[tb]
      \setlength{\sz}{.2\linewidth}
      \centering
      \scriptsize
      \resizebox{\linewidth}{!}{%
        \begin{tabular}{%
            lllll
          }
          (1) Correct & (2) Correct & (3) SG explainable & (4) SG explainable & (5) Objectively Correct \\
          \includegraphics[width=\sz]{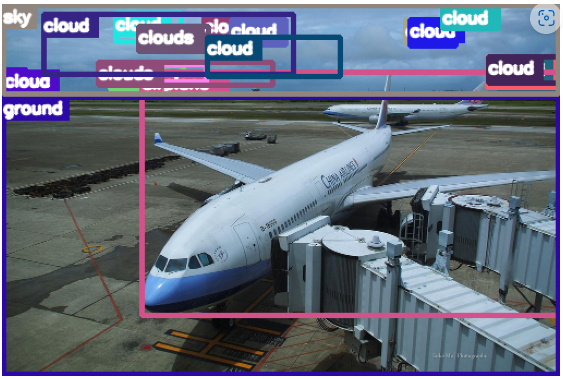} &
          \includegraphics[width=\sz]{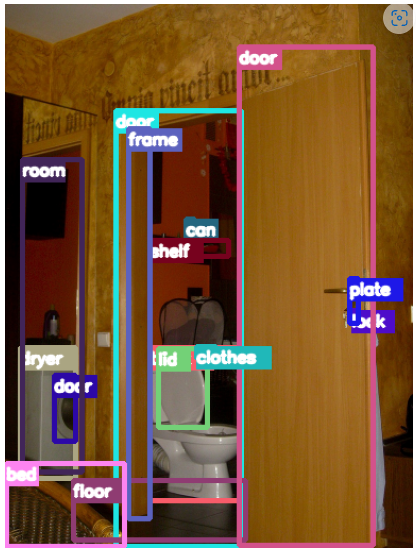} &
          \includegraphics[width=\sz]{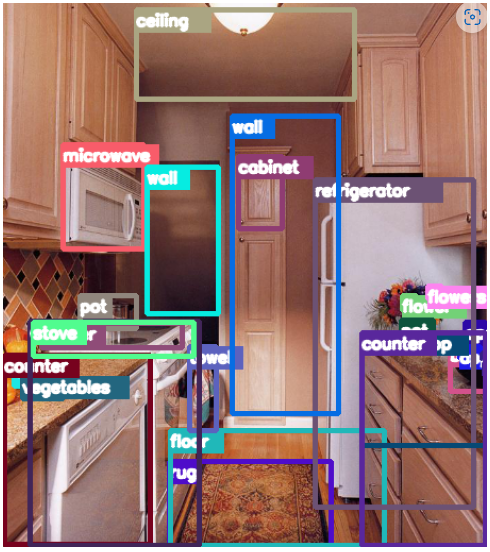} &
          \includegraphics[width=\sz]{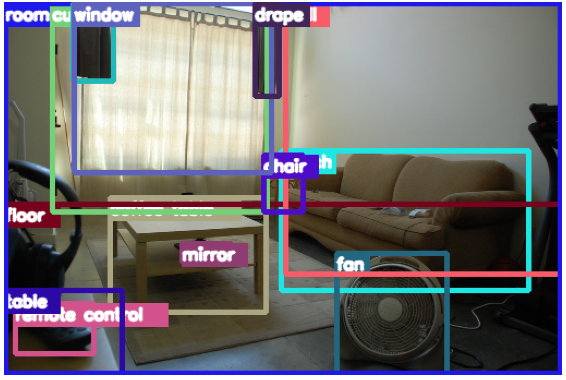} &
          \includegraphics[width=\sz]{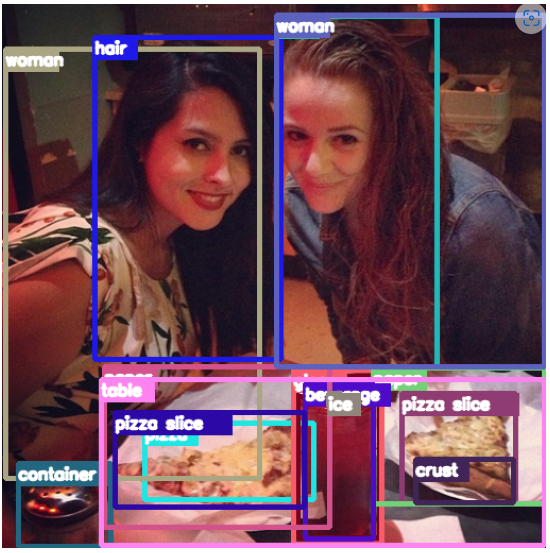} \\
          \includegraphics[width=\sz]{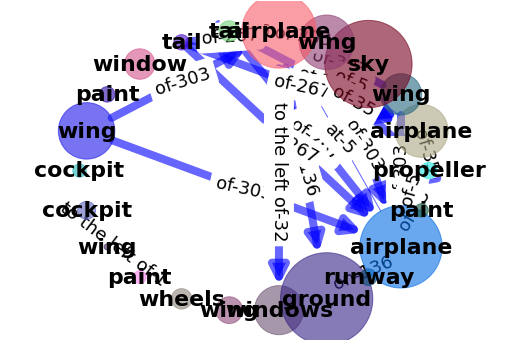} &
          \includegraphics[width=\sz]{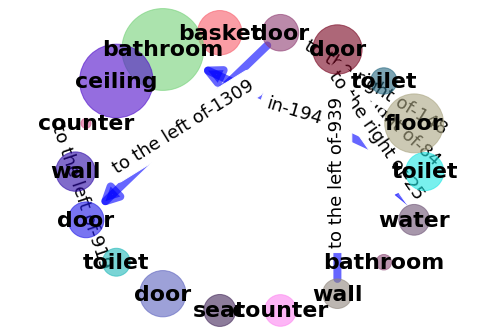} &
          \includegraphics[width=\sz]{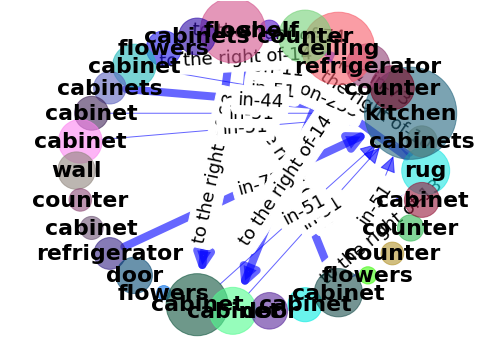} &
          \includegraphics[width=\sz]{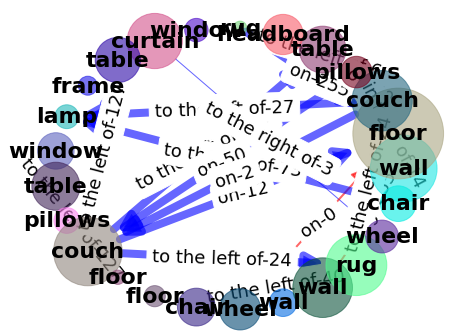} &
          \includegraphics[width=\sz]{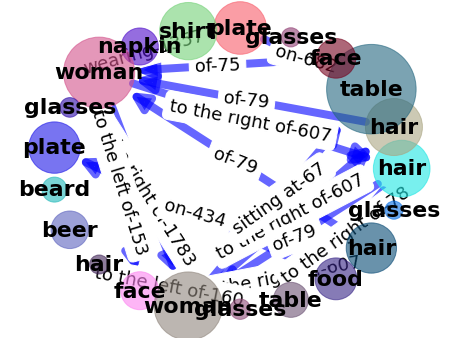} \\
          Q: What is the aircraft &
          Q: Are there any parachutes &
          Q: What is the white pot holding? &
          Q: Which kind of furniture &
          Q: What is in the red glass??\\
          on the ground? &
          or bags? &
          &
          is right of the curtains?
          \\
          Answer: Airplane &
          Answer: No &
          Answer: Flower  &
          Answer: Chair &
          Answer: Beverage\\
          Prediction: Airplane &
          Prediction: No &
          Prediction: Flowers &
          Prediction: Chair &
          Prediction: Liquid\\
        \end{tabular}
      }
      \caption{Examples to demonstrate the complexity of VQA and the explainability of the scene graph. All example is predicted by the SelfSim framework.}
      \label{fig:examples_app}
    \end{figure*}

    We present additional examples to illustrate how scene graphs can contribute to the explainability of AI in the context of VQA, Figure~\ref{fig:examples_app}.
    These examples highlight that VQA remains an open area of research and that the performance of a model should be evaluated beyond standard metrics.
    These examples serve as a reminder that there is room for further exploration and improvement in the field of VQA, extending beyond conventional evaluation metrics.

    All examples were predicted by the SelfSim framework.
    In the following discussion, the additional examples demonstrate both the problem of low agreement of VQA question answers due to ambiguity and the usefulness of scene graphs in providing more explainable AI for this task.

    For instance in example 1, the model accurately predicts the answer, and the detection of the airplane in the scene graph is easily visualized.
    Conversely, in example 2, the model correctly do not detect the object mentioned in the question, leading to a correct answer of 'No'.

    The benefits of using scene graphs for visual question answering become more evident in examples 3 and 4.
    In example 3, the model provides an objectively correct answer despite a different ground truth answer in the dataset.
    This discrepancy is explained by the scene graph, which highlights that the extracted object related to the question is 'flowers' rather than 'flowers'. In example 4, the model correctly classifies the link that relates the chair located to the right of the curtains in the scene graph, enabling the model to predict the correct answer.

    In example 5, the acceptance of the model's answer 'liquid' as opposed to the ground truth 'beverage' is subjective and depends on the evaluator's opinion.
    This demonstrates that the model's response may fail to precisely evaluate the question, emphasizing the inherent challenges in VQA.

    Overall, these examples highlight the potential benefits of incorporating scene graphs in visual question answering, offering insights into the model's reasoning and contributing to more interpretable AI systems.

